\documentclass[letterpaper, 10 pt, conference]{ieeeconf}  

\usepackage{cite}
\usepackage{amsmath,amssymb,amsfonts}
\usepackage{algorithmic}
\usepackage{graphicx}
\usepackage{textcomp}
\usepackage{multirow}
\usepackage[font=small]{caption}
\usepackage{subcaption}
\usepackage[table,xcdraw]{xcolor}
\usepackage{hyperref}
\usepackage{balance}
\usepackage{tcolorbox} 
\usepackage{xcolor} 
\IEEEoverridecommandlockouts                              

\title{\LARGE \bf
HapticVLM: VLM-Driven Texture Recognition Aimed at Intelligent Haptic Interaction
}

\author{Muhammad Haris Khan$^{*}$, Miguel Altamirano Cabrera$^{*}$, Dmitrii Iarchuk$^{*}$, \\Yara Mahmoud, Daria Trinitatova, Issatay Tokmurziyev, Dzmitry Tsetserukou
\thanks{*Equal Contribution}
\thanks{The authors are with the Intelligent Space Robotics Laboratory, Center for Digital Engineering, Skolkovo Institute of Science and Technology. 
       {\tt \{haris.khan, m.altamirano, Dmitrii.Iarchuk, Yara.Mahmoud, Daria.Trinitatova, issatay.tokmurziyev,  d.tsetserukou\}@skoltech.ru}}
}

\begin{document}

\maketitle
\thispagestyle{empty}
\pagestyle{empty}

\begin{abstract}

This paper introduces HapticVLM, a novel multimodal system that integrates vision-language reasoning with deep convolutional networks to enable real-time haptic feedback. HapticVLM leverages a ConvNeXt-based material recognition module to generate robust visual embeddings for accurate identification of object materials, while a state-of-the-art Vision-Language Model (Qwen2-VL-2B-Instruct) infers ambient temperature from environmental cues. The system synthesizes tactile sensations by delivering vibrotactile feedback through speakers and thermal cues via a Peltier module, thereby bridging the gap between visual perception and tactile experience. Experimental evaluations demonstrate an average recognition accuracy of 84.67\% across five distinct auditory-tactile patterns and a temperature estimation accuracy of 86.7\% based on a tolerance-based evaluation method with an 8°C margin of error across 15 scenarios. Although promising, the current study is limited by the use of a small set of prominent patterns and a modest participant pool. Future work will focus on expanding the range of tactile patterns and increasing user studies to further refine and validate the system's performance. Overall, HapticVLM presents a significant step toward context-aware, multimodal haptic interaction with potential applications in virtual reality, and assistive technologies.

\end{abstract}


\section{INTRODUCTION}

The ability to perceive and distinguish material properties such as texture, temperature, and stiffness is a fundamental aspect of human interaction with the physical world. Human tactile perception integrates visual, auditory, and haptic cues to form a comprehensive understanding of object surfaces, enabling precise material recognition and interaction \cite{lederman2009haptic}. While recent advances in computer vision and machine learning have significantly improved object detection \cite{7780460} and classification \cite{yu2205coca}, the replication of fine-grained tactile perception remains an open challenge. In particular, haptic feedback systems often rely on predefined material characteristics or direct tactile sensing, limiting their ability to adapt dynamically to new environments. With the emergence of Vision-Language Models (VLMs), deep learning (DL) has demonstrated an enhanced capability to infer object properties from multimodal inputs \cite{drehwald2023one}. However, the application of VLMs in haptic feedback systems has remained largely unexplored.

This paper presents HapticVLM, a novel system that leverages Convolutional Neural Network (CNN) for material recognition and VLM for environmental assessment to generate real-time haptic feedback.
The system operates through two primary stages. Initially, a neural network classifies the material of the object, namely metal, wood, or fabric, based on its visual characteristics. Upon classification, the system retrieves pre-recorded surface interaction sounds associated with the identified material and reproduces the corresponding haptic sensation using a speaker. By integrating vibration and auditory cues, HapticVLM enables users to perceive textures in a manner that closely simulates real-world tactile interactions. The second stage involves estimating the thermal properties of the object based on environmental conditions. Utilizing VLM-driven reasoning, the system analyzes contextual visual cues, including ambient lighting and object reflectivity, to infer the expected temperature. This estimated thermal response is rendered through a Peltier module, allowing users to experience dynamic thermal feedback that corresponds to the inferred material temperature.
\begin{figure}[t!]
  \centering
  \includegraphics[width=0.48\textwidth]{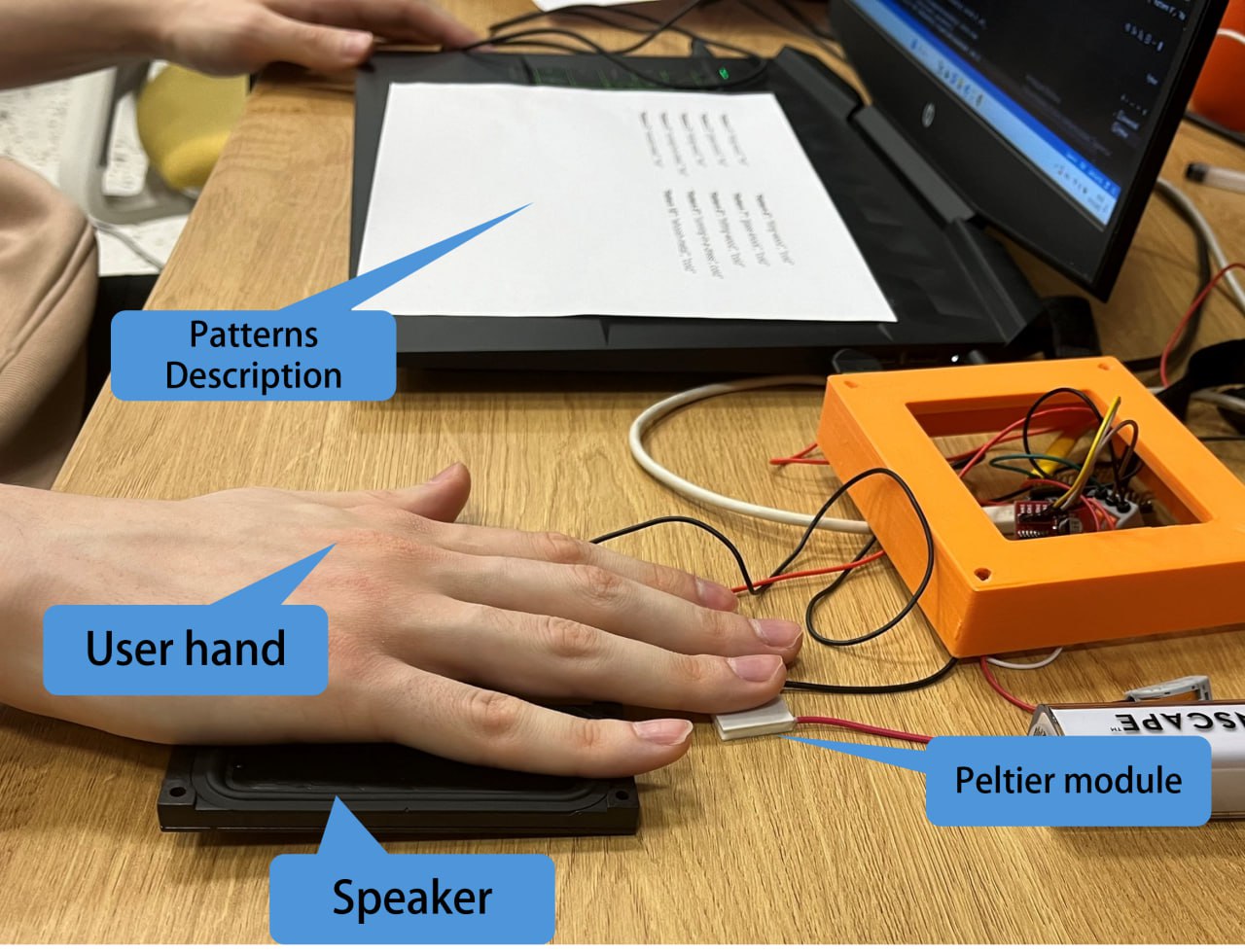}
  \caption{System overview of HapticVLM.}
  \label{fig:device}
\end{figure}

\begin{figure*}[t!]
    \centering
    \includegraphics[width=\textwidth]{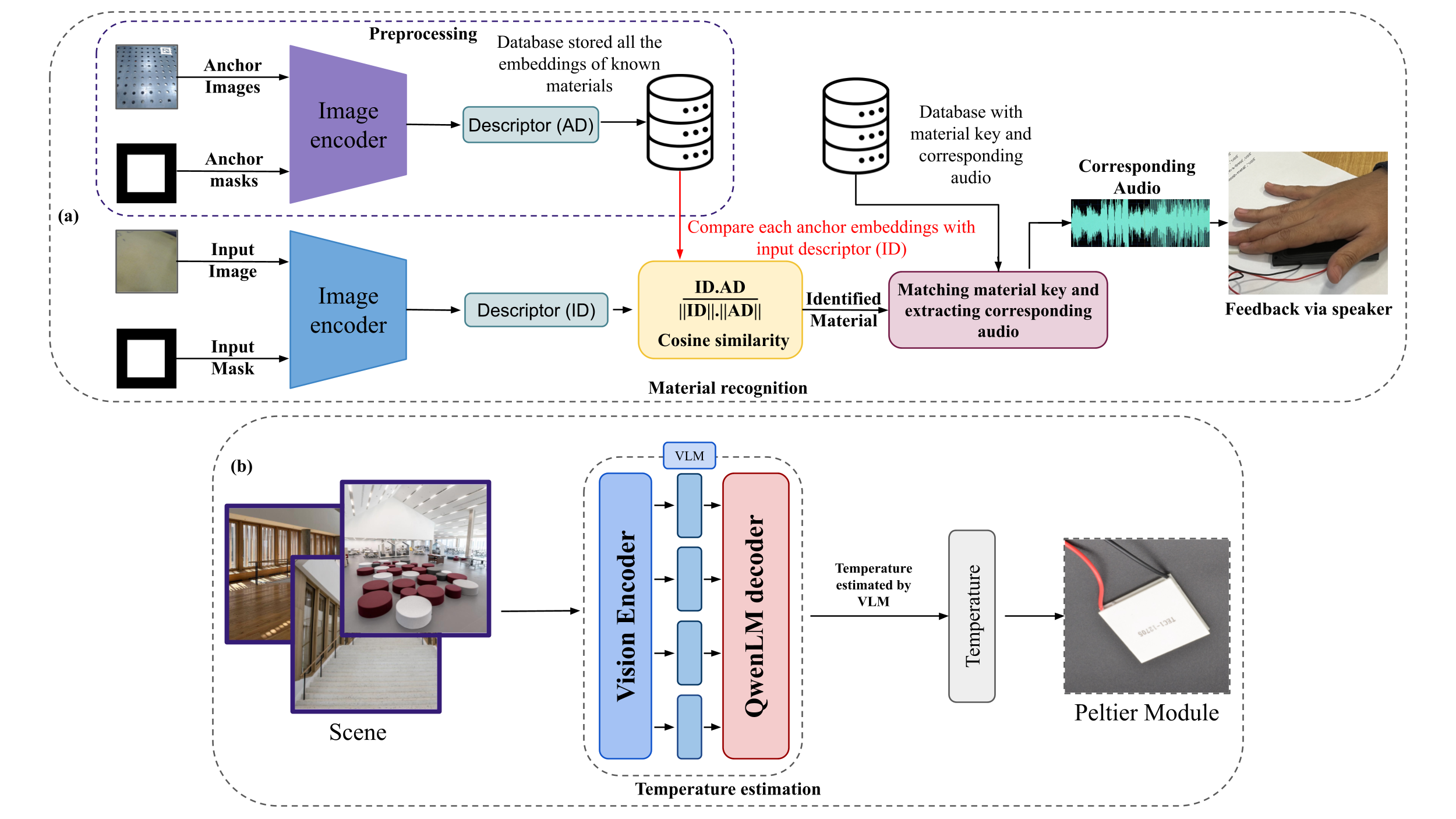} 
    \caption{Material recognition via ConvNeXT and temperature estimation via VLM. (a) The first step in material recognition involves preprocessing, which utilizes an image encoder to create embeddings of known materials stored in a database. The second step inputs images from the targeted scene into the same encoder to generate descriptors, followed by calculating cosine distance to identify the material. Once confirmed, the material name serves as a key to retrieve corresponding audio feedback from another database, which is then delivered to the user via a speaker. (b) Temperature estimation is performed using a Vision-Language Model (VLM), which considers visual cues such as clothing and lighting to estimate temperature, providing feedback to users through a Peltier device.
}
    \label{fig:system_architecture}
\end{figure*}
Unlike existing haptic feedback approaches, which rely on predefined parameters or handcrafted material properties, HapticVLM introduces a data-driven method that synthesizes haptic sensations dynamically. Prior research in haptic systems has explored vibration-based texture simulation \cite{bensmaia2003vibrations} and force-based kinesthetic feedback \cite{pacchierotti2017wearable}, yet these methods remain limited in their ability to integrate real-time perceptual reasoning. Our approach bridges vision, sound, and touch into a unified system by leveraging VLMs not only for object recognition but also for material property inference, which is subsequently converted into haptic feedback. Additionally, while previous systems have used fixed temperature values per material \cite{kim2020thermal}, HapticVLM differentiates by dynamically selecting from multiple predefined thermal states based on environmental cues, ensuring a more realistic simulation of temperature variations.

The implications of this work extend beyond virtual and augmented reality, with potential applications in assistive technologies, robotic teleoperation, and multisensory learning. For visually impaired individuals, the ability to "feel" objects through AI-driven haptic rendering could significantly enhance spatial awareness and object perception. In robotics, the integration of VLM-based tactile perception could enable manipulators to sense and convey material properties to human operators, improving teleoperation precision in hazardous environments. Additionally, educational applications could benefit from interactive learning experiences that incorporate realistic haptic cues for material science and engineering studies.

By combining multimodal perception with AI-driven haptic feedback, HapticVLM presents a new paradigm for intelligent, context-aware haptic interaction. The system establishes a foundation for next-generation haptic technologies, enabling richer, more immersive, and perceptually coherent experiences beyond traditional tactile feedback.

\section{Related Work}

Early haptic systems laid the groundwork for modern force-feedback and tactile interfaces. The PHANToM interface pioneered kinesthetic force feedback, enabling users to perceive virtual object geometry and stiffness by dynamically applying forces on a stylus \cite{salisbury2004haptic}. However, initial designs relied primarily on static force models, which could convey shape and hardness but struggled to render high-frequency details such as surface texture and friction.

Subsequent research addressed these limitations by integrating vibrotactile feedback, which introduced high- and low-frequency vibrations to improve realism. Altamirano et al. implemented vibromotors to analyze the tactile perception on the palm of the users while studing the interaction to different objects \cite{altamirano2020}   Culbertson et al. demonstrated that replaying prerecorded vibration signals could replicate coarse textures, significantly enhancing realism compared to purely force-based haptics \cite{culbertson2013generating}. In parallel, event-based and audio-driven approaches explored tool-surface interaction sounds as a means of generating vibrotactile signals, enabling more realistic tactile cues from auditory input \cite{sinapov2011interactive}, \cite{kuchenbecker2006improving}, \cite{cabrera2024}. This shift underscored a key principle in haptics: multimodal cues (sound, force, vibration) often enhance realism more effectively than a single modality.

As robotics and virtual reality advanced, researchers explored automatic material property inference using multiple sensing modalities. Vision-tactile fusion became a key research direction, where deep neural networks trained on both images and physical touch data improved object and material classification \cite{gao2016deep}. Additionally, high-resolution optical tactile sensors like GelSight enabled robots to capture detailed surface deformations, bridging the gap between vision-based and tactile sensing \cite{yuan2017gelsight}. However, such systems typically required direct physical contact and dense sensor arrays, limiting their applicability in purely virtual or remote environments. To address this, researchers investigated audio and vision as proxies for touch. Owens et al. showed that friction sounds and visual cues could help neural networks predict surface roughness and compliance \cite{owens2016visually}. Further studies explored the combination of proprioception, audio, and vision to classify objects, effectively training robots or machine-learning models to associate external cues with haptic outcomes \cite{sinapov2011interactive}, \cite{gao2016deep}. Despite these advances, real-time sensor-free haptic rendering—where specialized tactile hardware is not required at runtime—remains an open challenge.

Beyond force and vibration, thermal perception plays a critical role in simulating material conductivity (e.g., metals are felt as cool, plastics as neutral) and environmental context (ambient temperature). Early wearable thermoelectric devices enabled users to experience temperature variations in VR, but these systems often used predefined temperature profiles rather than real-time adaptive feedback \cite{peiris2017thermovr}. In robotics, infrared sensors and thermal imaging have been employed to estimate temperature and material properties \cite{erickson2020multimodal}, though such methods generally require specialized cameras or direct physical contact, making them impractical for remote or fully virtual environments. Recent research has explored learning emissivity and thermal conductivity from visual data, but real-time, generalizable solutions remain in early development.

Recent advances in Vision-Language Models (VLMs) have enabled the processing of multimodal inputs, including images, videos, and natural language—for improved decision-making. Transformer-based architectures such as BLIP-2 \cite{li2023blip}, Flamingo \cite{alayrac2022flamingo}, Kosmos-2 \cite{peng2023kosmos}, and Molmo and Pixmo \cite{deitke2024molmopixmoopenweights} have significantly enhanced the ability to interpret and integrate visual and textual information. Among these, Qwen2-VL \cite{wang2024qwen2} stands out as a state-of-the-art open-source model that offers robust generalization and efficient multimodal learning, making it highly suitable for haptic applications requiring real-time reasoning.

Parallel to these advancements, researchers have explored personalized haptic feedback to account for individual differences in force, thermal, and vibrotactile sensitivity \cite{a10010015}. This is especially crucial in assistive technologies, where visual information is translated into vibrotactile or auditory cues for users with visual impairments \cite{kristjansson2016designing}. However, material-specific haptic feedback, such as distinguishing between stone and cardboard textures, remains challenging, since many assistive systems rely on coarse heuristics rather than context-aware inferences.

Temporal synchronization across multiple haptic cues (e.g., sound, vibration, temperature) plays a crucial role in perceived realism. Neural network-based methods have been explored to align these modalities, though they typically rely on well-curated datasets \cite{culbertson2013generating}, \cite{sinapov2011interactive}. Neuromorphic tactile sensing has emerged as a promising approach to improve responsiveness while reducing power consumption in real-time applications \cite{rasouli2018extreme}, though its adoption is currently hindered by specialized manufacturing requirements.

Given these developments, a key missing piece is real-time, sensor-free haptic rendering that adaptively fuses visual context (e.g., surface reflectivity, scene lighting), language cues (e.g., material labels, semantic descriptors), and user preferences into a unified pipeline. Current research lacks vision-language material recognition models that generate multimodal haptic feedback (vibrotactile + thermal) without requiring physical sensors or pre-existing material libraries. 
By leveraging large-scale VLMs to infer environmental conditions and utilizing CNNs to recognize material properties, a system could dynamically generate vibrotactile and thermal feedback in real time. 
This opens the door to scalable, context-aware haptics for applications in VR, telepresence, robotics, and assistive technologies.

\section{System Architecture}
\subsection{Material Recognition}

Figure \ref{fig:system_architecture}a illustrates the pipeline for material recognition module of HapticVLM system. The methodology proposed in \cite{drehwald2023one} serves as the foundation for this task. This paper outlines a comprehensive framework for identifying and distinguishing various materials in images, demonstrating effectiveness across multiple applications. The ConvNeXt architecture is employed to generate embeddings for materials. This DL model is specifically designed to extract high-quality feature representations from images, ensuring that the embeddings accurately capture the unique characteristics of each material.
The input data for the model consists of an image containing the material alongside a mask indicating its location within the image. The image acts as the primary source of visual information, while the mask highlights the region of interest, allowing the model to focus on relevant features. To construct a robust database for recognition tasks, we selected
N materials and preprocessed their embeddings. These materials were chosen to ensure diversity and relevance, with their embeddings computed and stored after preprocessing.
Real-time recognition of materials from video is achievable, meaning that as video frames are processed sequentially, the system can swiftly identify materials in each frame without significant delays. For these images, the same mask is applied under the assumption that the material is centrally located within each frame. This simplification posits that the material of interest occupies a prominent position, thereby reducing computational complexity while maintaining accuracy.
Subsequently, the resulting embedding is compared with those in the database using cosine similarity (1). By calculating cosine similarity between embeddings, the system efficiently determines how closely a given material matches those stored in the database, facilitating accurate recognition even in challenging scenarios.
This approach integrates advanced DL techniques with practical applications, enhancing material recognition capabilities across various contexts.

\begin{equation}
\cos(\theta) = \frac{\mathbf{A} \cdot \mathbf{B}}{\|\mathbf{A}\| \|\mathbf{B}\|} = \frac{\sum_{i=1}^{n} A_i B_i}{\sqrt{\sum_{i=1}^{n} A_i^2}\,\sqrt{\sum_{i=1}^{n} B_i^2}}
\tag{1}
\end{equation}

\begin{figure}[hb]
\centerline{\includegraphics[width=0.5\textwidth]{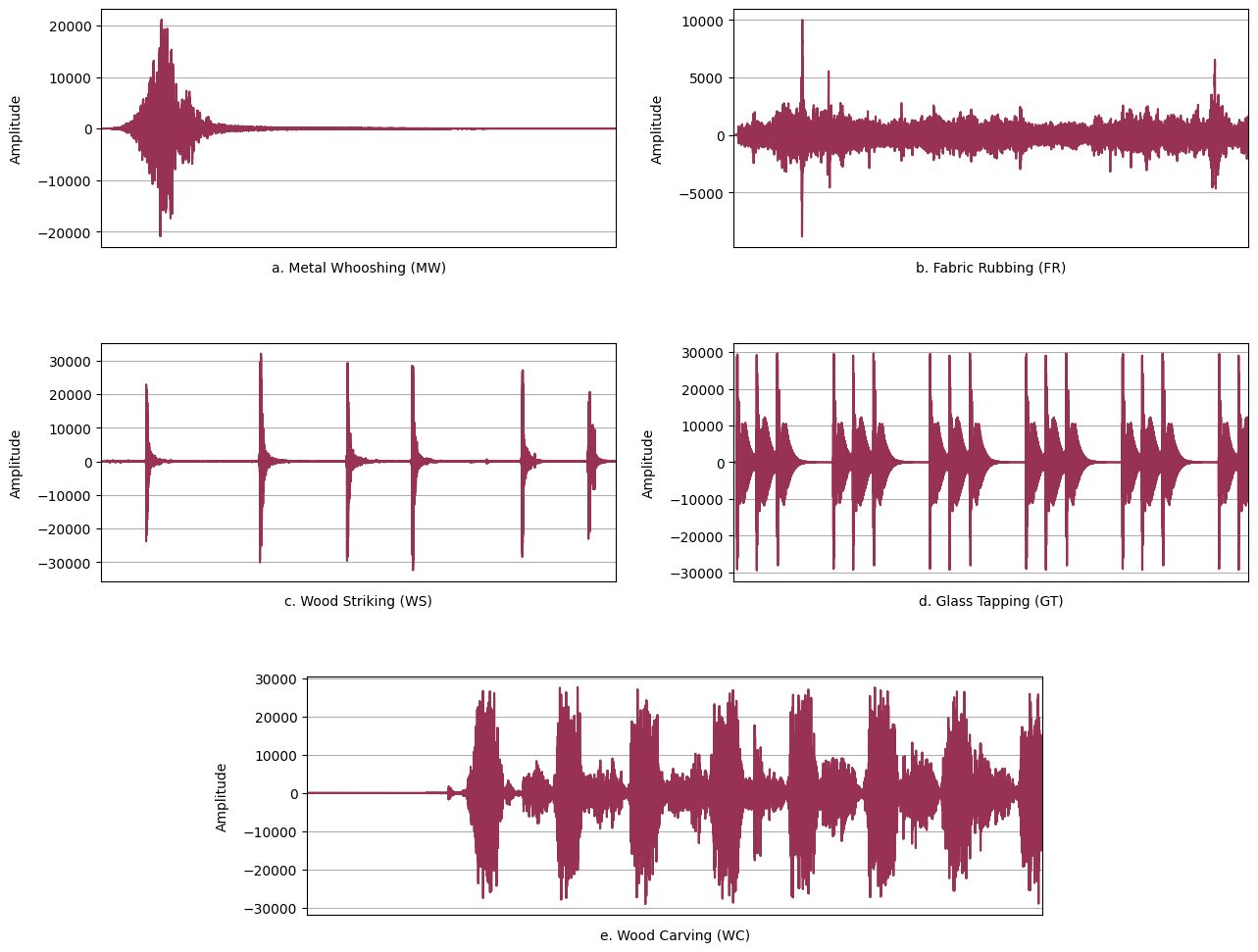}}
\caption{Haptic feedback patterns (a) Metal Whooshing  (MW), (b) Fabric Rubbing (FR), (c) Wood striking (WS), (d) Glass tapping (GT), and (e) Wood carving (WC)}
\label{fig:haptics}
\vspace{-0.4cm}
\end{figure}

\begin{figure}[t]
\centerline{\includegraphics[width=0.5\textwidth]{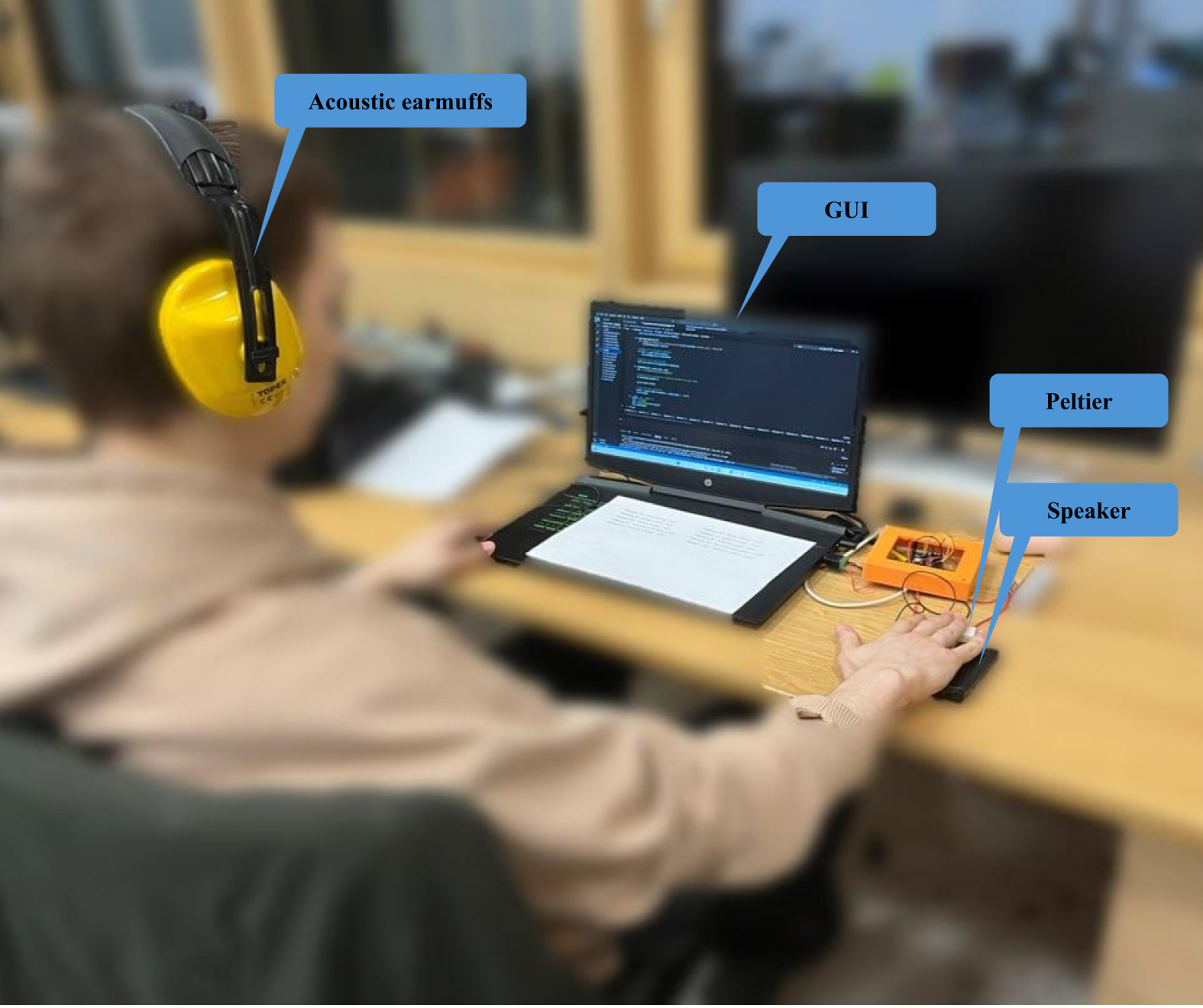}}
\caption{Participant seated at a desk during the evaluation, with their right hand placed on the device for haptic feedback assessment.}
\label{fig:haptics1}
\vspace{-0.4cm}
\end{figure}

\subsection{VLM-based Temperature Inference}

Figure \ref{fig:system_architecture}b illustrates the temperature estimation task. To determine the room temperature, a Vision-Language Model (VLM) was employed, which processes an image captured by the camera of the surrounding space along with a query regarding the temperature in the room based on the photo. We utilize Qwen2-VL-2B-Instruct \cite{wang2024qwen2}, an open-source model that excels in generalization and efficient multimodal learning, making it particularly suitable for applications requiring real-time reasoning.
This approach leverages the strengths of VLMs to interpret visual cues effectively, enabling accurate temperature estimation based on contextual information derived from the image. The integration of visual data with natural language queries allows for a more intuitive interaction, enhancing user experience by providing immediate feedback based on environmental conditions.
Such capabilities are critical for developing responsive systems that can adapt to varying scenarios and deliver reliable information to users in real time. The use of VLMs in this context exemplifies their versatility and potential in practical applications across different domains.

\subsection{Haptic Feedback Module}

Speakers were utilized to provide physical interaction through vibrations. Users can feel these vibrations by placing their palms on the speakers. The sounds reproduced are carefully selected to closely resemble actual tactile sensations. It is important to note that humans can physically perceive vibrations primarily for sounds with frequencies ranging from 1 to 1000 hertz \cite{shi2020evaluation}, which imposes a limitation on audio selection.
The integration of speakers in this manner enhances the sensory experience, allowing users to connect auditory and tactile feedback effectively. This approach is particularly beneficial in applications such as gaming, virtual reality, and immersive environments, where realistic interactions are crucial. By ensuring that the audio output aligns with the tactile feedback, the system creates a more engaging and intuitive user experience.

For temperature feedback, we utilized a Peltier module (TEC1-03108, 20×20$mm$). This thermoelectric device operates by creating a temperature differential when an electric current is passed through it, allowing one side to become hot while the other side cools down. The Peltier module is particularly effective for applications requiring precise temperature control, as it can respond rapidly to changes in operating conditions.

\begin{table*}[]
\vspace{3mm}
\centering{
\caption{Confusion Matrix for Actual and Perceived Pattern Recognition.}
\label{table:confusion} 
\setlength{\tabcolsep}{6pt} 
\renewcommand{\arraystretch}{1} 

\begin{tabular}{|cc|cccccccccc|}
\hline
\multicolumn{2}{|c|}{}                                   & \multicolumn{10}{c|}{Answers   (Predicted Class)}                                                                                                                                                                                                                                                                                                                                                                                                                                                                                                                                                                                                                                                                                                      \\ \cline{3-12} 
\multicolumn{2}{|c|}{\multirow{-2}{*}{\%}}               & \multicolumn{1}{c|}{WC-h}                                                & \multicolumn{1}{c|}{GT-h}                                                & \multicolumn{1}{c|}{WS-h}                                                & \multicolumn{1}{c|}{FR-h}                                                & \multicolumn{1}{c|}{MW-h}                                                & \multicolumn{1}{c|}{WC-c}                                                & \multicolumn{1}{c|}{GT-c}                                                & \multicolumn{1}{c|}{WS-c}                                                & \multicolumn{1}{c|}{FR-c}                                                & MW-c                                                \\ \hline
\multicolumn{1}{|c|}{}                            & WC-h & \multicolumn{1}{c|}{\cellcolor[HTML]{0B3040}{\color[HTML]{FFFFFF} 1.00}} & \multicolumn{1}{c|}{\cellcolor[HTML]{FFFFFF}-}                        & \multicolumn{1}{c|}{\cellcolor[HTML]{FFFFFF}-}                        & \multicolumn{1}{c|}{\cellcolor[HTML]{FFFFFF}-}                        & \multicolumn{1}{c|}{\cellcolor[HTML]{FFFFFF}-}                        & \multicolumn{1}{c|}{\cellcolor[HTML]{FFFFFF}-}                        & \multicolumn{1}{c|}{\cellcolor[HTML]{FFFFFF}-}                        & \multicolumn{1}{c|}{\cellcolor[HTML]{FFFFFF}-}                        & \multicolumn{1}{c|}{\cellcolor[HTML]{FFFFFF}-}                        & \cellcolor[HTML]{FFFFFF}-                        \\ \cline{2-12} 
\multicolumn{1}{|c|}{}                            & GT-h & \multicolumn{1}{c|}{\cellcolor[HTML]{FFFFFF}-}                        & \multicolumn{1}{c|}{\cellcolor[HTML]{1C3E4D}{\color[HTML]{FFFFFF} 0.93}} & \multicolumn{1}{c|}{\cellcolor[HTML]{EFF2F3}0.07}                        & \multicolumn{1}{c|}{\cellcolor[HTML]{FFFFFF}-}                        & \multicolumn{1}{c|}{\cellcolor[HTML]{FFFFFF}-}                        & \multicolumn{1}{c|}{\cellcolor[HTML]{FFFFFF}-}                        & \multicolumn{1}{c|}{\cellcolor[HTML]{FFFFFF}-}                        & \multicolumn{1}{c|}{\cellcolor[HTML]{FFFFFF}-}                        & \multicolumn{1}{c|}{\cellcolor[HTML]{FFFFFF}-}                        & \cellcolor[HTML]{FFFFFF}-                        \\ \cline{2-12} 
\multicolumn{1}{|c|}{}                            & WS-h & \multicolumn{1}{c|}{\cellcolor[HTML]{FFFFFF}-}                        & \multicolumn{1}{c|}{\cellcolor[HTML]{F5F6F7}0.04}                        & \multicolumn{1}{c|}{\cellcolor[HTML]{375562}{\color[HTML]{FFFFFF} 0.82}} & \multicolumn{1}{c|}{\cellcolor[HTML]{EFF2F3}0.07}                        & \multicolumn{1}{c|}{\cellcolor[HTML]{FFFFFF}-}                        & \multicolumn{1}{c|}{\cellcolor[HTML]{FFFFFF}-}                        & \multicolumn{1}{c|}{\cellcolor[HTML]{FFFFFF}-}                        & \multicolumn{1}{c|}{\cellcolor[HTML]{FAFBFB}0.02}                        & \multicolumn{1}{c|}{\cellcolor[HTML]{FFFFFF}-}                        & \cellcolor[HTML]{F5F6F7}0.04                        \\ \cline{2-12} 
\multicolumn{1}{|c|}{}                            & FR-h & \multicolumn{1}{c|}{\cellcolor[HTML]{FFFFFF}-}                        & \multicolumn{1}{c|}{\cellcolor[HTML]{FFFFFF}-}                        & \multicolumn{1}{c|}{\cellcolor[HTML]{FAFBFB}0.02}                        & \multicolumn{1}{c|}{\cellcolor[HTML]{3C5A67}{\color[HTML]{FFFFFF} 0.80}} & \multicolumn{1}{c|}{\cellcolor[HTML]{DADFE2}0.16}                        & \multicolumn{1}{c|}{\cellcolor[HTML]{FFFFFF}-}                        & \multicolumn{1}{c|}{\cellcolor[HTML]{FFFFFF}-}                        & \multicolumn{1}{c|}{\cellcolor[HTML]{FFFFFF}-}                        & \multicolumn{1}{c|}{\cellcolor[HTML]{FAFBFB}0.02}                        & \cellcolor[HTML]{FFFFFF}-                        \\ \cline{2-12} 
\multicolumn{1}{|c|}{}                            & MW-h & \multicolumn{1}{c|}{\cellcolor[HTML]{EAEDEF}0.09}                        & \multicolumn{1}{c|}{\cellcolor[HTML]{FFFFFF}-}                        & \multicolumn{1}{c|}{\cellcolor[HTML]{FFFFFF}-}                        & \multicolumn{1}{c|}{\cellcolor[HTML]{EFF2F3}0.07}                        & \multicolumn{1}{c|}{\cellcolor[HTML]{3C5A67}{\color[HTML]{FFFFFF} 0.80}} & \multicolumn{1}{c|}{\cellcolor[HTML]{FFFFFF}-}                        & \multicolumn{1}{c|}{\cellcolor[HTML]{FFFFFF}-}                        & \multicolumn{1}{c|}{\cellcolor[HTML]{FAFBFB}0.02}                        & \multicolumn{1}{c|}{\cellcolor[HTML]{FAFBFB}0.02}                        & \cellcolor[HTML]{FFFFFF}-                        \\ \cline{2-12} 
\multicolumn{1}{|c|}{}                            & WC-c & \multicolumn{1}{c|}{\cellcolor[HTML]{EFF2F3}0.07}                        & \multicolumn{1}{c|}{\cellcolor[HTML]{FFFFFF}-}                        & \multicolumn{1}{c|}{\cellcolor[HTML]{FFFFFF}-}                        & \multicolumn{1}{c|}{\cellcolor[HTML]{FFFFFF}-}                        & \multicolumn{1}{c|}{\cellcolor[HTML]{FFFFFF}-}                        & \multicolumn{1}{c|}{\cellcolor[HTML]{1C3E4D}{\color[HTML]{FFFFFF} 0.93}} & \multicolumn{1}{c|}{\cellcolor[HTML]{FFFFFF}-}                        & \multicolumn{1}{c|}{\cellcolor[HTML]{FFFFFF}-}                        & \multicolumn{1}{c|}{\cellcolor[HTML]{FFFFFF}-}                        & \cellcolor[HTML]{FFFFFF}-                        \\ \cline{2-12} 
\multicolumn{1}{|c|}{}                            & GT-c & \multicolumn{1}{c|}{\cellcolor[HTML]{FFFFFF}-}                        & \multicolumn{1}{c|}{\cellcolor[HTML]{F5F6F7}0.04}                        & \multicolumn{1}{c|}{\cellcolor[HTML]{FFFFFF}-}                        & \multicolumn{1}{c|}{\cellcolor[HTML]{FFFFFF}-}                        & \multicolumn{1}{c|}{\cellcolor[HTML]{FFFFFF}-}                        & \multicolumn{1}{c|}{\cellcolor[HTML]{FAFBFB}0.02}                        & \multicolumn{1}{c|}{\cellcolor[HTML]{3C5A67}{\color[HTML]{FFFFFF} 0.80}} & \multicolumn{1}{c|}{\cellcolor[HTML]{DFE4E6}0.13}                        & \multicolumn{1}{c|}{\cellcolor[HTML]{FFFFFF}-}                        & \cellcolor[HTML]{FFFFFF}-                        \\ \cline{2-12} 
\multicolumn{1}{|c|}{}                            & WS-c & \multicolumn{1}{c|}{\cellcolor[HTML]{FFFFFF}-}                        & \multicolumn{1}{c|}{\cellcolor[HTML]{FFFFFF}-}                        & \multicolumn{1}{c|}{\cellcolor[HTML]{EAEDEF}0.09}                        & \multicolumn{1}{c|}{\cellcolor[HTML]{FAFBFB}0.02}                        & \multicolumn{1}{c|}{\cellcolor[HTML]{FFFFFF}-}                        & \multicolumn{1}{c|}{\cellcolor[HTML]{FFFFFF}-}                        & \multicolumn{1}{c|}{\cellcolor[HTML]{FFFFFF}-}                        & \multicolumn{1}{c|}{\cellcolor[HTML]{425E6B}{\color[HTML]{FFFFFF} 0.78}} & \multicolumn{1}{c|}{\cellcolor[HTML]{EFF2F3}0.07}                        & \cellcolor[HTML]{F5F6F7}0.04                        \\ \cline{2-12} 
\multicolumn{1}{|c|}{}                            & FR-c & \multicolumn{1}{c|}{\cellcolor[HTML]{FFFFFF}-}                        & \multicolumn{1}{c|}{\cellcolor[HTML]{FFFFFF}-}                        & \multicolumn{1}{c|}{\cellcolor[HTML]{FFFFFF}-}                        & \multicolumn{1}{c|}{\cellcolor[HTML]{FAFBFB}0.02}                        & \multicolumn{1}{c|}{\cellcolor[HTML]{FFFFFF}-}                        & \multicolumn{1}{c|}{\cellcolor[HTML]{FFFFFF}-}                        & \multicolumn{1}{c|}{\cellcolor[HTML]{FAFBFB}0.02}                        & \multicolumn{1}{c|}{\cellcolor[HTML]{EFF2F3}0.07}                        & \multicolumn{1}{c|}{\cellcolor[HTML]{47636F}{\color[HTML]{FFFFFF} 0.76}} & \cellcolor[HTML]{DFE4E6}0.13                        \\ \cline{2-12} 
\multicolumn{1}{|c|}{\multirow{-10}{*}{\rotatebox{90}{Patterns}}} & MW-c & \multicolumn{1}{c|}{\cellcolor[HTML]{FFFFFF}-}                        & \multicolumn{1}{c|}{\cellcolor[HTML]{FFFFFF}-}                        & \multicolumn{1}{c|}{\cellcolor[HTML]{FFFFFF}-}                        & \multicolumn{1}{c|}{\cellcolor[HTML]{FFFFFF}-}                        & \multicolumn{1}{c|}{\cellcolor[HTML]{F5F6F7}0.04}                        & \multicolumn{1}{c|}{\cellcolor[HTML]{FFFFFF}-}                        & \multicolumn{1}{c|}{\cellcolor[HTML]{FFFFFF}-}                        & \multicolumn{1}{c|}{\cellcolor[HTML]{FAFBFB}0.02}                        & \multicolumn{1}{c|}{\cellcolor[HTML]{EAEDEF}0.09}                        & \cellcolor[HTML]{31515E}{\color[HTML]{FFFFFF} 0.84} \\ \hline

\end{tabular}}
\end{table*}

\section{Experimental Evaluation}

\subsection{VLM-driven Temperature Estimation}
To assess the ability of the applied VLM to accurately estimate scene temperatures, we conducted an experiment using 15 images with known temperature values.
For each image, we calculated the absolute error between the VLM's predicted temperature and the actual temperature. A prediction was deemed correct if this error was less than or equal to 8°C. We selected an 8°C tolerance as it provides a logical balance: it is broad enough to accommodate minor, acceptable deviations due to the inherent uncertainty in visual cues, yet sufficiently strict to flag predictions that are meaningfully off target. Out of the 15 images, 13 had predictions that fell within the 8°C tolerance, resulting in an overall accuracy of approximately 86.7\%. This indicates that in nearly 87\% of cases, the VLM’s temperature predictions were close to the actual values, demonstrating robust performance. Nonetheless, the two images with errors of 10°C and 12°C suggest that there is still room for improvement. Future efforts may focus on refining model training, integrating additional contextual cues, and expanding the dataset to further enhance accuracy.


\subsection{Haptic Pattern Recognition Study}

To assess the effectiveness of the proposed haptic feedback module to generate  recognizable tactile patterns, a user study was conducted. Five distinct vibration stimuli were selected to simulate common material interactions: wood carving (WC), glass tapping (GT), wood striking (WS), fabric rubbing (FR), and metal whooshing (MW), as shown in Fig.~\ref{fig:haptics}. Additionally, two thermal conditions, hot (h) and cold (c), were incorporated using a Peltier module. Consequently, a total of ten unique tactile patterns were generated and presented to the participants for evaluation.

\subsubsection{Experimental Setup}
Nine participants (7 males, 2 females, aged 22–35 years, mean 26 $\pm$3.9) completed the study. After providing informed consent, participants underwent a training session to familiarize themselves with the patterns. Each pattern was rendered three times during training, and a visual reference of the patterns was provided throughout the session.

During the evaluation, participants were asked to sit in front of a desk and to locate their right hand on the device, as shown in Fig.~\ref{fig:haptics1}. The experimenter used a graphical user interface (GUI) on a PC to select the patterns that the users perceived. Each of the 10 patterns (five sounds, two thermal conditions) was presented five times in random order, resulting in 50 trials per participant. Participants provided feedback on the perceived sensations at the end of the study.

\subsubsection{Results}
Recognition of tactile patterns averaged 84.7\%, with the highest recognition rate achieved for the wood carving hot (WC-h) pattern (100\%) and the lowest for the fabric rubbing cold (FR-c) pattern (75.6\%). A confusion matrix summarizing the results is shown in Table~\ref{table:confusion}.

A two-way repeated measures ANOVA was conducted to assess the impact of vibration pattern and temperature on recognition accuracy. The analysis revealed no statistically significant main effect for vibration (F(9, 72) = 1.92, p = 0.063) or temperature (F(1, 8) = 2.59, p = 0.146). Furthermore, the interaction between vibration and temperature was not significant (F(9, 72) = 1.05, p = 0.410), indicating that recognition accuracy did not significantly differ across the various vibrations or their thermal classifications.

The effect sizes were calculated using partial eta squared ($np^2$) to provide insight into the variance explained by each factor. The $np^2$ for pattern was 0.193428, suggesting a medium effect size, while temperature yielded an $np^2$ of 0.244444, indicating a moderate effect size. The interaction effect had a smaller effect size $np^2$ = 0.115974).

To further investigate potential differences between individual patterns, pairwise comparisons were performed using paired t-tests. Despite some comparisons showing T-statistics suggestive of differences, none reached statistical significance at the conventional alpha level of 0.05 after applying a Bonferroni correction (all corrected p-values = 1.000).

The obtained results indicate that participants exhibited a consistent level of recognition accuracy across the tested patterns, regardless of whether they were classified as hot or cold. This finding suggests that the specific patterns used in this experiment did not produce significant differences in recognition capabilities among the participants.

\section{Conclusions and Future Work}
In this paper, we presented HapticVLM, a novel system that leverages Vision-Language Models (VLMs) and deep convolutional networks for generating real-time, multimodal haptic feedback. Our system architecture comprises two primary modules: material recognition and temperature estimation. The material recognition module employs a ConvNeXt-based encoder to generate embeddings from material images, facilitating robust identification through cosine similarity measures. Concurrently, the temperature estimation module utilizes the Qwen2-VL-2B-Instruct model to infer ambient temperature from visual cues, which is then rendered through a Peltier module. Auditory cues synchronized with tactile vibrations further enhance the multisensory experience provided by the system.

Experimental evaluations demonstrated promising results in both domains. The haptic pattern recognition study achieved an average accuracy of 84.7\%, with distinct patterns such as wood carving hot (WC-h) achieving a perfect recognition rate. In the temperature estimation task, the VLM correctly inferred the temperature range in 13 out of 15 cases, corresponding to an accuracy of 86.7\%. These results highlight the system’s potential in dynamically synthesizing realistic haptic feedback based on visual and auditory inputs.

Nevertheless, our current study is subject to several limitations. The haptic pattern recognition experiment was conducted using only five prominent vibrotactile patterns, which may not fully capture the variability encountered in real-world scenarios. Moreover, the participant pool was relatively small, limiting the generalizability of our findings.

In future work, we plan to expand the experimental design by incorporating a broader array of tactile patterns that are more similar in nature, to better understand the nuances in user perception and system performance. Additionally, we intend to increase the number of participants to obtain more statistically robust insights. Further research will explore the integration of additional sensory modalities, such as force feedback, and the refinement of the VLM-based temperature estimation through advanced model training and larger, more diverse datasets. These improvements aim to enhance the realism and adaptability of the haptic feedback system for applications in virtual reality, teleoperation, and assistive technologies.

Overall, HapticVLM represents a significant step toward intelligent, context-aware haptic interaction, bridging the gap between visual perception and tactile sensation, and setting the stage for more immersive and effective multisensory systems in the future.



\end{document}